%% file: naaclhlt2019.tex
\documentclass[11pt,a4paper]{article}
\usepackage[hyperref]{naaclhlt2019}
\usepackage{times}
\usepackage{latexsym}
\usepackage{tikz}
\usepackage{todonotes}

\usepackage{url}
\usepackage{graphicx}

\aclfinalcopy % Uncomment this line for the final submission
\title{Dialogue Act Classification with Context-Aware Self-Attention}

\author{Vipul Raheja \hspace{0.3cm} Joel Tetreault\\
  Grammarly \\
  {\tt firstname.lastname@grammarly.com} \\}

\date{}

\begin{document}
\maketitle
\begin{abstract}
Recent work in Dialogue Act classification has treated the task as a sequence labeling problem using hierarchical deep neural networks. We build on this prior work by leveraging the effectiveness of a context-aware self-attention mechanism coupled with a hierarchical recurrent neural network. We conduct extensive evaluations on standard Dialogue Act %\todo{make sure to spell/capitalize consistently!} 
classification datasets and show significant improvement over state-of-the-art results on the Switchboard Dialogue Act (SwDA) Corpus. We also investigate the impact of different utterance-level representation learning methods and show that our method is effective at capturing utterance-level semantic text representations while maintaining high accuracy.
\end{abstract}

\section{Introduction}

Dialogue Acts (DAs) are the functions of utterances in dialogue-based interaction \cite{austin1975things}. A DA represents the meaning of an utterance at the level of illocutionary force, and hence, constitutes the basic unit of linguistic communication \cite{searle1969speech}. DA classification is an important task in Natural Language Understanding, with applications in question answering, conversational agents, speech recognition, etc. Examples of DAs can be found in Table \ref{corpus-examples}.  Here we have a conversation of 7 utterances between two speakers.  Each utterance has a corresponding label such as {\it Question} or {\it Backchannel}.

Early work in this field made use of statistical machine learning methods %(HMMs, SVMs, CRFs) 
and approached the task as either a structured prediction or text classification problem \cite{stolcke2000dialogue, ang2005automatic, zimmermann2009joint, surendran2006dialog}. 
Many recent studies have proposed deep learning models for the DA classification task with promising results \cite{lee2016sequential,khanpour2016dialogue,ortega2017neural}. However, most of these approaches treat the task as a text classification problem, treating each utterance in isolation, rendering them unable to leverage the conversation-level contextual dependence among utterances. Knowing the text and/or the DA labels of the previous utterances can assist in predicting the current DA state. For instance, in Table \ref{corpus-examples}, the \textit{Answer} or \textit{Statement} dialog acts often follow \textit{Question} type utterances.
%This can be observed in , which shows a series of utterances and their corresponding labels 
%snippet of a conversation with utterances and their corresponding labels 

\begin{table}[t!]
\begin{center}
\small
\begin{tabular}{llllll}
\hline \bf Speaker & \bf Utterance & \bf DA label \\  \hline 
A & Okay. & Other\\
A & Um, what did you do this \\
& weekend? & Question \\
B & Well, uh, pretty much spent\\
& most of my time in the yard. & Statement\\
B & [Throat Clearing] & Non Verbal\\
A & Uh-Huh. & Backchannel\\
A & What do you have planned \\
& for your yard? & Question\\
B & Well, we're in the process\\
& of, revitalizing it. & Statement\\ 
\hline
\end{tabular}
\end{center}
\caption{\label{corpus-examples}  A snippet of a conversation sample from the SwDA Corpus. Each utterance has a corresponding dialogue act label.}
\end{table}

%Our work develops a dialogue-act classification model that is effective across multiple domains, by drawing on recent advances in NLP such as self-attention.  
%\todo{are there any other advances we can list here?  Or just leave it as "recent advances in NLP."  I guess the self-attention component is the big novelty that differentiates us from the prior art?}
%\todo{VR: Changed to previous version of this sentence}
% including self-attention, hierarchical deep learning models, and contextual dependencies. 
This work draws from recent advances in NLP such as self-attention, hierarchical deep learning models, and contextual dependencies to produce a dialogue act classification model that is effective across multiple domains.  
Specifically, we propose a hierarchical deep neural network to model different levels of utterance and dialogue act semantics, achieving state-of-the-art performance on the Switchboard Dialogue Act Corpus. We demonstrate how performance can improve by leveraging context at different levels of the model: previous labels for sequence prediction (using a CRF), conversation-level context with self-attention for utterance representation learning, and character embeddings at the word-level. Finally, we explore different ways to learn effective utterance representations, which serve as the building blocks of our hierarchical architecture for DA classification.  
%\todo{CN: thinks that can be added to intro: first self-attention for DAC; a conclusion from utterance representation learning experiments}

%% JRT - old end paragraph
%We draw ideas dispersed across these prior works, using a hierarchical deep neural network to model different levels of utterance and dialogue act semantics, achieving state-of-the-art performance on the Switchboard Dialogue Act Corpus. We demonstrate how leveraging context at different levels of the model: previous labels (using a CRF) for sequence prediction, conversation-level context with self-attention for utterance representation learning, and character-embeddings at the word-level, can improve model performance. Finally, we explore different ways to learn effective utterance representations, which serve as the building blocks of our hierarchical architecture for DA classification.  

%of a hierarchical RNN architecture to classify its dialogueact in the context of the rest of the conversation.
% The contributions of this work are (i) Proposing a novel hierarchical neural network architecture for Dialogue Sequence Labeling, leveraging context at multiple levels in the hierarchy. (ii) Giving empirical evaluations of this model on benchmark data sets, and analyzing different ways of learning utterance representations. (iii) Achieving state-of-the-art performance with this end-to-end system.
%\todo{are 2 and 3 kind of the same thing?}

\section{Related Work}
\label{ssec:related-work}
A full review of all DA classification methods is outside the scope of the paper, thus we focus on two main classes of approaches which have dominated recent research: those that treat DA classification as a text classification problem, where each utterance is classified in isolation, and those that treat it as a  sequence labeling problem. 
%\citep{lee2016sequential,khanpour2016dialogue,ji2016latent,shen2016neural,ortega2017neural}, or as a sequence labeling problem \citep{kalchbrenner2013recurrent,li2016multi,AAAI1816706,chen2017dialogue,tran2017hierarchical,liu2017using}. 

\noindent {\bf Text Classification}: \newcite{lee2016sequential} build a vector representation for each utterance, using either a CNN or RNN, and use the preceding utterance(s) as context to classify it. Their model was extended by \newcite{khanpour2016dialogue} and \newcite{ortega2017neural}. \newcite{shen2016neural}  used a variant of the attention-based encoder for the task. \newcite{ji2016latent} use a hybrid architecture, combining an RNN language model with a latent variable model.% for joint modeling of utterance and DA labels.

\noindent {\bf Sequence Labeling}: \newcite{kalchbrenner2013recurrent} used a mixture of sentence-level CNNs and discourse-level RNNS to achieve state-of-the-art results on the task. %\newcite{lee2016sequential} built a vector representation for each utterance, using either a CNN or RNN, and use the representation of the preceding utterance(s) as context to classify it with a feed-forward network. 
% \todo[inline]{\begin{enumerate} \item you don't need the contrastive here \item an example in the non-textual features might be nice (e.g., features, such as \dots \item shallow discourse structure? \end{enumerate}}
% \todo{\begin{enumerate}\item Not sure what you mean by \textit{neural attention model} here as you focus on the architectures for the rest. In other words, is the new thing that they add attention to their models? \item meaning the rest were not applied to DA?\end{enumerate} }
%\newcite{khanpour2016dialogue} used a deep LSTM-based RNN architecture using pre-trained word embeddings similar to the RNN-based model of \newcite{lee2016sequential}.
% \todo{%
% 	In my opinion, you should cluster them by topic (or method) not chronologically or by author. That'll make your life easier \textit{and} the paragraph more readable (you won't have to pay attention to ten different names). Here's what I understand from the above:
% 	\begin{enumerate}
% 		\item \newcite{kalchbrenner2013recurrent} sentence - hierarchical CNN - discourse RNN
% 		\item \newcite{lee2016sequential}  CNN + RNN classification based on the prev
% 		\item \newcite{ji2016latent} RNN + LVM
% 		\item \newcite{li2016multi} Hierarchical RNN + non-textual
% 		\item \newcite{shen2016neural} Neural attention model (whatever that means)
% 		\item \newcite{khanpour2016dialogue} Deep RNN + pre-trained
% 		\item \newcite{ortega2017neural} CNN --> RNN attention
% 		\item \newcite{liu2017using} CNN $\rightarrow$ (CNN|LSTM) $\rightarrow$ DA
%     \end{enumerate}
% }
Recent works \cite{li2016multi, liu2017using} have increasingly employed hierarchical architectures to learn and model multiple levels of utterance and DA dependencies. 
%developed a hierarchical model to learn utterance representations using CNNs, and used CNN or LSTM to model the higher-level DA sequences. 
% \newcite{li2016multi} proposed a hierarchical RNN model combining RNN-based utterance representations with non-textual features. 
\newcite{AAAI1816706}, \newcite{chen2017dialogue} and \newcite{tran2017hierarchical} used RNN-based hierarchical neural networks, using different combinations of techniques like last-pooling or attention mechanism to encode sentences, coupled with CRF decoders.  \newcite{chen2017dialogue} achieved the highest performance to date on the two datasets for this task.

% More recently, \newcite{kumar2017dialogue},  \newcite{chen2017dialogue} and \newcite{tran2017hierarchical} employed hierarchical RNNs, using techniques like last-pooling, or attention mechanism to encode sentences, coupled with CRF decoders, to learn and model multiple levels of utterance and dialogue act dependencies.\todo{did all three do the above or parts of it?} \todo{Different parts of it, what I described was the union} \todo{Dimi: for me it's not really clear (a) who did what/how are the papers different, and (b) from the way it is written how yours \textit{really} relates to only the three above. Hierarchical RNNs were also done in Li and Wu (apparently).} 
Our work extends these hierarchical models and leverages a combination of techniques proposed across these prior works (CRF decoding, contextual attention, and character-level word embeddings) with self-attentive representation learning, and is able to achieve state-of-the-art performance. 

\section{Model}
The task of DA classification
%\todo{I just realized that nowhere in this paper do we give examples of DA labels.  Do other short papers in the field do this? -- Not really}
takes a conversation $C$ as input, which is a varying length sequence of utterances $U = \{u_1, u_2, ... u_{L}\}$. Each utterance $u_i \in U$, in turn, is a sequence of varying lengths of words  $\{w_{i}^{1}, w_{i}^{2}, ..., w_{i}^{N_{i}}\}$, and has a corresponding target label $y_i \in Y$. Hence, each conversation (i.e. a sequence of utterances) is mapped to a corresponding sequence of target labels $Y = \{y_1, y_2, ..., y_{L}\}$, which represents the DAs associated with the corresponding utterances.

\begin{figure}
    \centering
    \def\svgwidth{\columnwidth}
    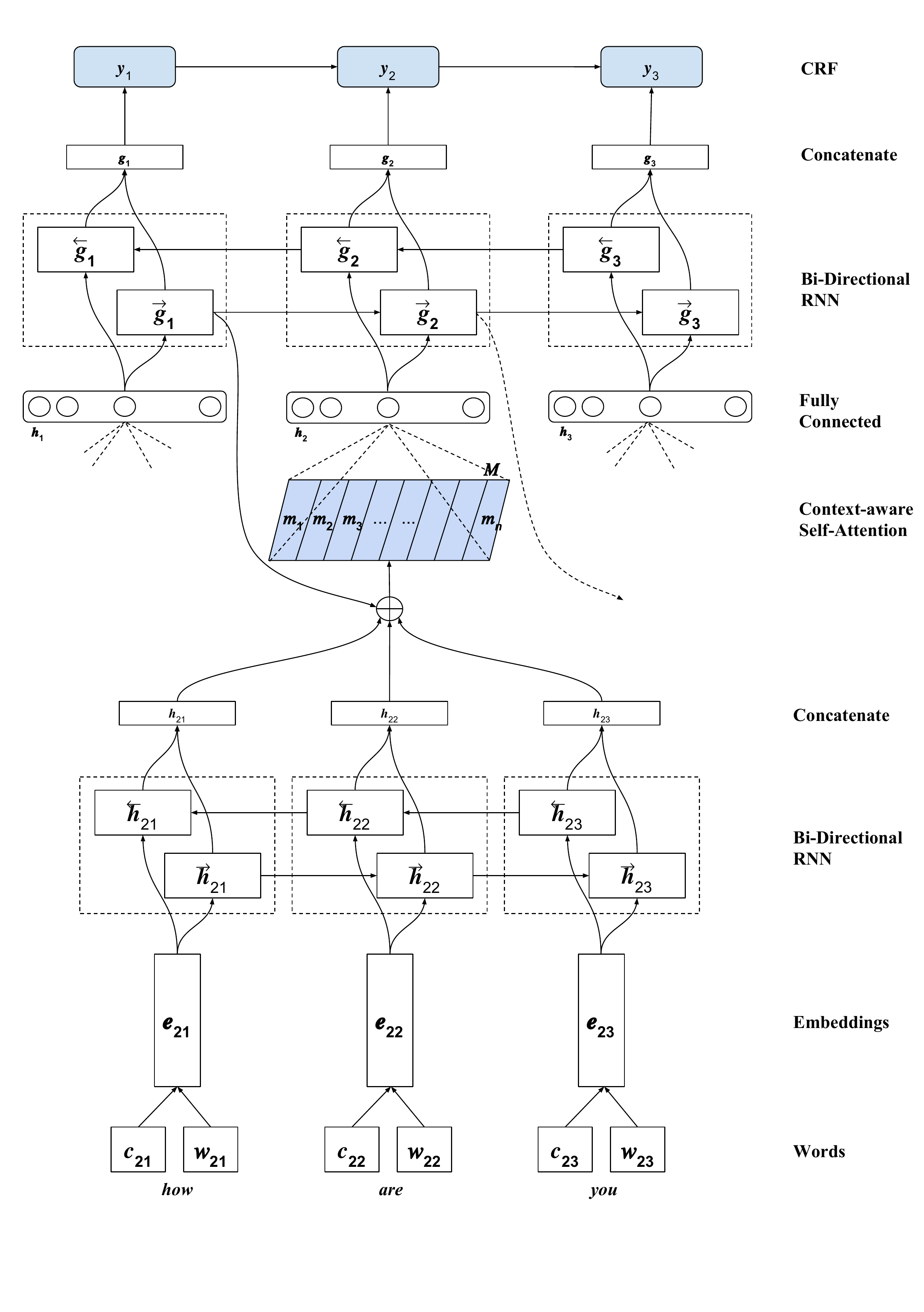
    \newline
    \newline
    \newline
    \caption{\label{figure1} Model Architecture}
\end{figure}

%\todo{commented out the "Why hierarchical model" paragraph, since it is partly covered in the Intro}
% The hierarchical architecture of the model, like much of the prior work, and the input itself, \todo{where was the data described? : I meant the input $C_i$, $U_i$ etc.} derives from the intuition about conversational structure that 
% %First, since conversations have an intrinsic hierarchical structure: words, composed by constituency and dependency relations, \todo{not sure if this is right?} form utterances; sequences of back-and-forth utterances then form a conversation. 
% contextual information at different levels in the conversation can be better leveraged by modeling dialogue utterances hierarchically. 
% Any other non-hierarchical arrangement would have to compromise on the benefits of either the utterance structure or conversational structure and context across utterances. Hence, we first build representations of utterances using words and then aggregate those at the conversation level.

Figure \ref{figure1}  shows the overall architecture of our model, which involves three main components: (1) an utterance-level RNN that encodes the information within the utterances at the word and character-level; (2) a context-aware self-attention mechanism that aggregates word representations into utterance representations; and (3) a conversation-level RNN that operates on the utterance encoding output of the attention mechanism, followed by a CRF layer to predict utterance labels. We describe them in detail  below.

\subsection{Utterance-level RNN}
For each word in an utterance, we combine two different word embeddings: GloVe \cite{pennington2014glove} and pre-trained ELMo representations \cite{peters2018deep} with fine-tuned task-specific parameters, which have shown superior performance in a wide range of tasks. The word embedding is then concatenated with its CNN-based 50-$D$ character-level embedding \cite{chiu2016named,ma2016end} to get the complete word-level representations. The motivation behind incorporating subword-level information is to infer the lexical features of utterances and named entities better. 

The word representation layer is followed by a bidirectional GRU (Bi-GRU) layer. %where the representation of each word at a time-step is obtained by concatenating the outputs from the forward and backward RNNs. 
Concatenating the forward and backward outputs of the Bi-GRU generates the utterance embedding that serves as input to the utterance-level context-aware self-attention mechanism which learns the final utterance representation. 

\subsection{Context-aware Self-attention}
Self-attentive representations encode a variable-length sequence into a fixed size, using an attention mechanism that considers different positions within the sequence. 
%Self-attention has been effectively applied not only to learn generic sentence representations, but also across a diverse range of tasks including author attribution, text entailment and sentiment classification \cite{?}.
Inspired by \newcite{tran2017hierarchical}, we use the previous hidden state from the conversation-level RNN (Section \ref{ssec:conversation-rnn}), which provides the context of the conversation so far, and combine it with the hidden states of all the constituent words in an utterance, into a self-attentive encoder \cite{linal2017embediclr}, which computes a $2D$ representation of each input utterance. We follow the notation originally presented in \newcite{linal2017embediclr} to explain our modification of their self-attentive sentence representation below.

An utterance $u_i$, which is a sequence of $n$ words $\{w_i^1, w_i^2, ... w_i^n\}$, is mapped into an embedding layer, resulting in a $d$-dimensional word embedding for every word. 
It is then fed into a bidirectional-GRU layer, whose hidden state outputs are concatenated at every time step.  

\begin{equation}
\overrightarrow{h_i^j} = \overrightarrow{GRU}(w_i^j,\overrightarrow{h_i^{j-1}})
\end{equation}
\begin{equation}
\overleftarrow{h_i^j} = \overleftarrow{GRU}(w_i^j,\overleftarrow{h_i^{j+1}})
\end{equation}
\begin{equation}
\mathbf{h_i^j} = concat(\overrightarrow{h_i^j},\overleftarrow{h_i^j})
\end{equation}
\begin{equation}
H_i = \{\mathbf{h_i^1}, \mathbf{h_i^2}, ... \mathbf{h_i^n}\}
\end{equation}

$H_i$ represents the $n$ GRU outputs of size $2u$ ($u$ is the number of hidden units in a unidirectional GRU). 

The contextual self-attention scores are then computed as follows: 
\begin{equation}
\label{attn-context-eqn}
S_i = W_{s2} tanh(W_{s1}H_i^T + {W_{s3}} \overrightarrow{g_{i-1}} + \mathbf{b})
\end{equation}
Here, $W_{s1}$ is a weight matrix with a shape of $d_a\times2u$, $W_{s2}$ is a matrix of parameters of shape $r \times d_a$, where $r$ and $d_a$ are hyperparameters we can set arbitrarily, and $W_{s3}$ is a parameter matrix of shape $d_a \times k$ for the conversational context, where $k$ is another hyperparameter that is the size of a hidden state in the conversation-level RNN (size of $\overrightarrow{g_{i-1}}$), and $\mathbf{b}$ is a vector representing bias. Equation \ref{attn-context-eqn} can then be treated as a 2-layer MLP with bias, with $d_a$ hidden units, ${W_{s1}, W_{s2}}$ and $W_{s3}$ as weight parameters. 
The scores $S_i$ are mapped into a probability matrix $A_i$ by means of a softmax function: 
\begin{equation}
A_i = softmax(S_i)
\end{equation}
which is then used to obtain a 2-d representation $M_i$ of the input utterance, using the GRU hidden states $H_i$ according to the attention weights provided by $A_i$ as follows: 
\begin{equation}
M_i = A_i H_i
\end{equation}
This 2-d representation is then projected to a 1-d embedding (denoted as $\mathbf{h_i}$), using a fully-connected layer. The conversation-level GRU then operates over this 1-d utterance embedding, and hence, we can represent $\mathbf{g_i}$ as:
\begin{equation}
\overrightarrow{g_i} = \overrightarrow{GRU}(\mathbf{h_i}, \overrightarrow{g_{i-1}})
\end{equation}
\begin{equation}
\overleftarrow{g_i} = \overrightarrow{GRU}(\mathbf{h_i}, \overrightarrow{g_{i+1}})
\end{equation}
\begin{equation}
\mathbf{g_i} = concat(\overrightarrow{g_i}, \overleftarrow{g_i})
\end{equation}
$\mathbf{g_i}$ then provides the conversation-level context used to learn the attention scores and 2-d representation ($M_{i+1}$) for the next utterance in the conversation ($\mathbf{h_{i+1}}$).

%%%%%%%%%%%%% OLD SECTION %%%%%%%%%%

% Formally, we compute the attention probability matrix as follows: 

% \begin{equation}
% A = g(W_{s_{2}} f(W_{s_{1}}H^T + {W_{s_{3}}} \overrightarrow{g_{t-1}} + \mathbf{b}))
% \end{equation}
% %\todo{the gt-1 plus b term is in bold, is this on purpose?} %\todo{Yes, bold represents vectors, and non-bold represents matrices.} 
% where $H$ represents the GRU hidden vectors at the utterance-level; $W_{s_{1}}$, $W_{s_{2}}$ and $W_{s_{3}}$ are trainable parameter matrices; $\overrightarrow{g_{t-1}}$ is the forward GRU hidden state of the previous utterance in the conversation; and $\mathbf{b}$ is a vector representing bias. The GRU hidden states $H$ are then summed up according to the weights provided by $A$ to get a $2D$ representation $M$ of the input utterance:

% \begin{equation}\label{utr-rep-eqn} M = AH \end{equation}

% This $2D$ representation is then projected to a $1D$ embedding using a fully-connected dense layer. For a more detailed explanation of the above, please refer to Section \ref{attention-appendix} in the Appendix.

%%%%%%%% OLD SECTION END %%%%%%%%%%%%%

\subsection{Conversation-level RNN}
\label{ssec:conversation-rnn}
The utterance representation $\mathbf{h_i}$ from the previous step is passed on to the conversation-level RNN, which is another bidirectional GRU layer used to encode utterances across a conversation. The hidden states  
\overrightarrow{g_i} and \overleftarrow{g_i} (Figure \ref{figure1}) are then concatenated to get the final representation $\mathbf{g_i}$ of each utterance, which is further propagated to a linear chain CRF layer. The CRF layer considers the correlations between labels in context and jointly decodes the optimal sequence of utterance labels for a given conversation, instead of decoding each label independently. 
%\todo{}

%%%%%%%%%%%%%%%%%%%%%%%%%%%%%%%%%%%%%%%%%%%%%%%%%%%%%%%%%%%%%%%%%%%%%%%%%%%%%%
%
% DATA
%

\section{Data}
We evaluate the classification accuracy of our model on the two standard datasets used for DA classification:  the Switchboard Dialogue Act Corpus (SwDA) \cite{jurafsky1997switchboard} consisting of 43 classes, and the 5-class version of the ICSI Meeting Recorder Dialogue Act Corpus (MRDA) introduced in \cite{ang2005automatic}.  For both datasets, we use the train, validation and test splits as defined in \newcite{lee2016sequential}. %\todo{probably could remove the foonote since you have the paper cited} %\footnote{\url{https://github.com/Franck-Dernoncourt/naacl2016}} 

\begin{table}[t!]
\begin{center}
\small
\begin{tabular}{llllll}
\hline \bf Dataset & \bf \textbar C\textbar & \bf \textbar V\textbar & \bf Train & \bf Validation & \bf Test \\ \hline
MRDA & 5 & 12k & 78k & 16k & 15k\\
SwDA & 43 & 20k & 193k & 23k & 5k\\
\hline
\end{tabular}
\end{center}
\caption{\label{font-table3} Number of utterances by dataset. \textbar C\textbar{} denotes number of classes and \textbar V\textbar{} is the vocabulary size.}
\vspace{-0.2cm}
\end{table}

Table {\ref{font-table3}} shows the statistics for both datasets.
%\todo{it's not immediately obvious what C and V are in the table} 
They are highly imbalanced in terms of class distribution, with the DA classes \texttt{Statement-non-opinion} and \texttt{Acknowledge/Backchannel} in SwDA and \texttt{Statement} in MRDA making up over 50\% of the labels in each set.

%with one class (\texttt{s}) in MRDA, and two classes (\texttt{sd} and \texttt{b}) in SwDA making up over 50\% of the label distribution.%\todo{we could add something about what code was used}
%%%%%%%%%%%%%%%%%%%%%%%%%%%%%%%%%%%%%%%%%%%%%%%%%%%%%%%%%%%%%%%%%%%%%%%%%%%%%%
%
% RESULTS
%
\section{Results}

\subsection{Dialogue Act Classification}

We compare the classification accuracy of our model against several other recent methods  (Table {\ref{da-classification-results}}).\footnote{Contemporaneous to this submission, \cite{li2018dual, wan2018improved, ravi2018self} proposed different approaches for the task. We do not focus on them here per NAACL 2019 guidelines, however note that our system outperforms the first two.  \cite{ravi2018self} bypasses the need for complex networks with huge parameters but its overall accuracy is 4.2\% behind our system, despite being 0.2\% higher on SwDA.}  
% CN reworded footnote
% \footnote{Very recently, \citealt{li2018dual, wan2018improved, ravi2018self} proposed different approaches for the task, and we would like to acknowledge their works despite their contemporaneous nature (they were made public right before our work was submitted) and thus we do not focus on them here give the NAACL19 guidelines.  That being said, the first two do not surpass the performance of our system.  \citealt{ravi2018self} is promising as it bypasses the need for complex networks with huge parameters.  However, it is overall 4.2\% behind our system in accuracy, despite being 0.2\% better on SwDA performance.}  
%of their works, we would like to point out that these works do not surpass the performance of our system.}
%\footnote{Very recently, \cite{ravi2018self} published their work at EMNLP 2018, which is quite promising as it bypasses the need for complex networks with huge parameters, it is overall 4.2\% behind our system in accuracy, despite being 0.2\% better on SwDA performance. While we acknowledge their work, we would like to point out that it came out only a few days before our work was submitted.} 
% As described in Section \ref{ssec:related-work}, all the other approaches use some form of a deep learning architecture. 
Four approaches \cite{chen2017dialogue,tran2017hierarchical,ortega2017neural,shen2016neural} use attention in some form to model the conversations, but none of them have explored self-attention for the task. The last three use CRFs in the final layer of sequence labeling. Only one other method \cite{chen2017dialogue} uses character-level word embeddings. All models and their variants were trained ten times and we report the average test performance. Our model outperforms state-of-the-art methods by 1.6\% on SwDA, the primary dataset for this task, and comes within 0.6\% on MRDA.  It also beats a TF-IDF GloVe baseline (described in Section \ref{utt-rep-learning-section}) by 16.4\% and 12.2\%, respectively. 

\begin{figure}
    % \vspace{-0.2cm}
    \centering
    \def\svgwidth{\columnwidth}
    \includegraphics[width=7cm]{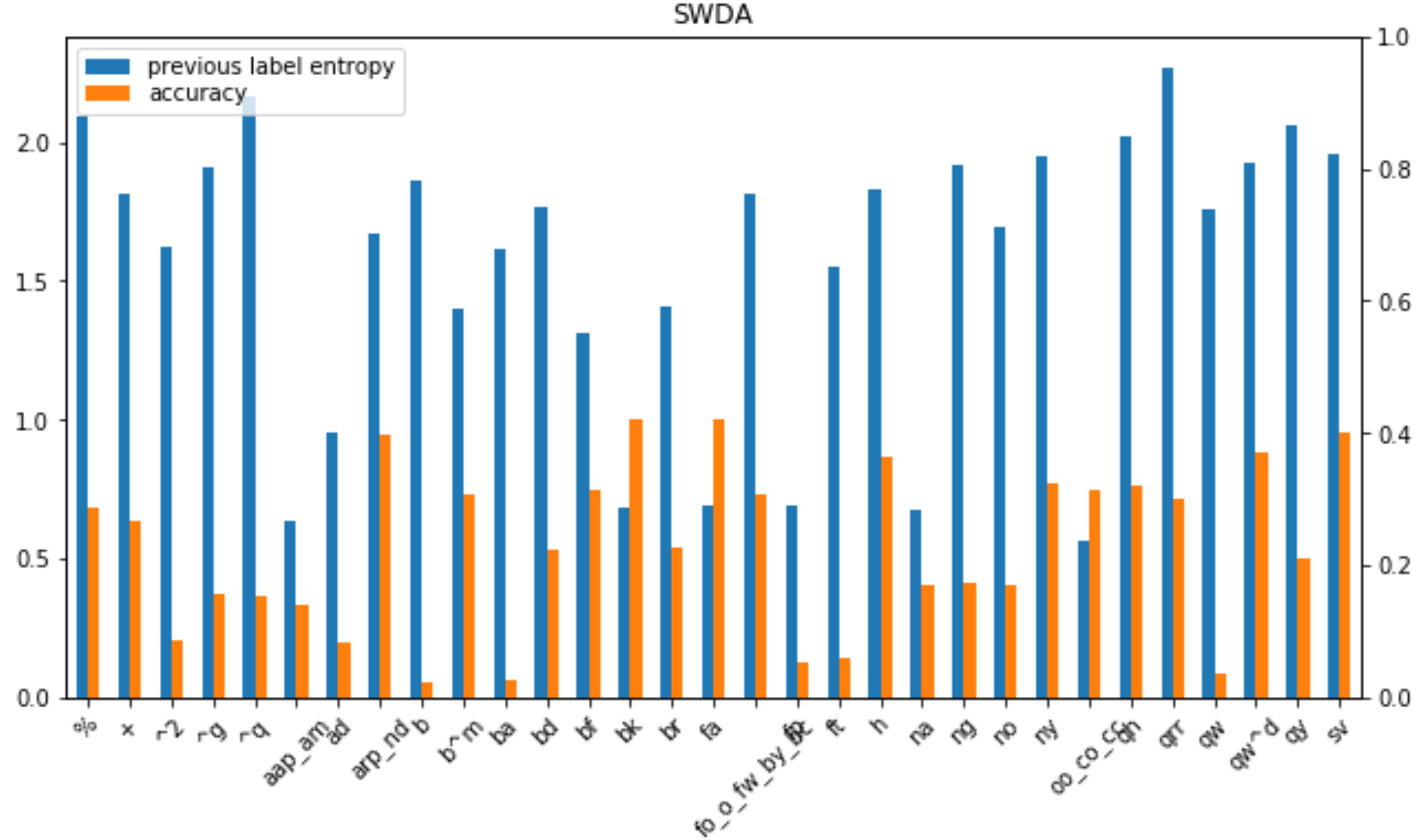}
    \caption{\label{figure2} Previous Label Entropy vs. Accuracy on the SwDA Dataset}
\end{figure}

\begin{figure}
    \centering
    \def\svgwidth{\columnwidth}
    \includegraphics[width=7cm]{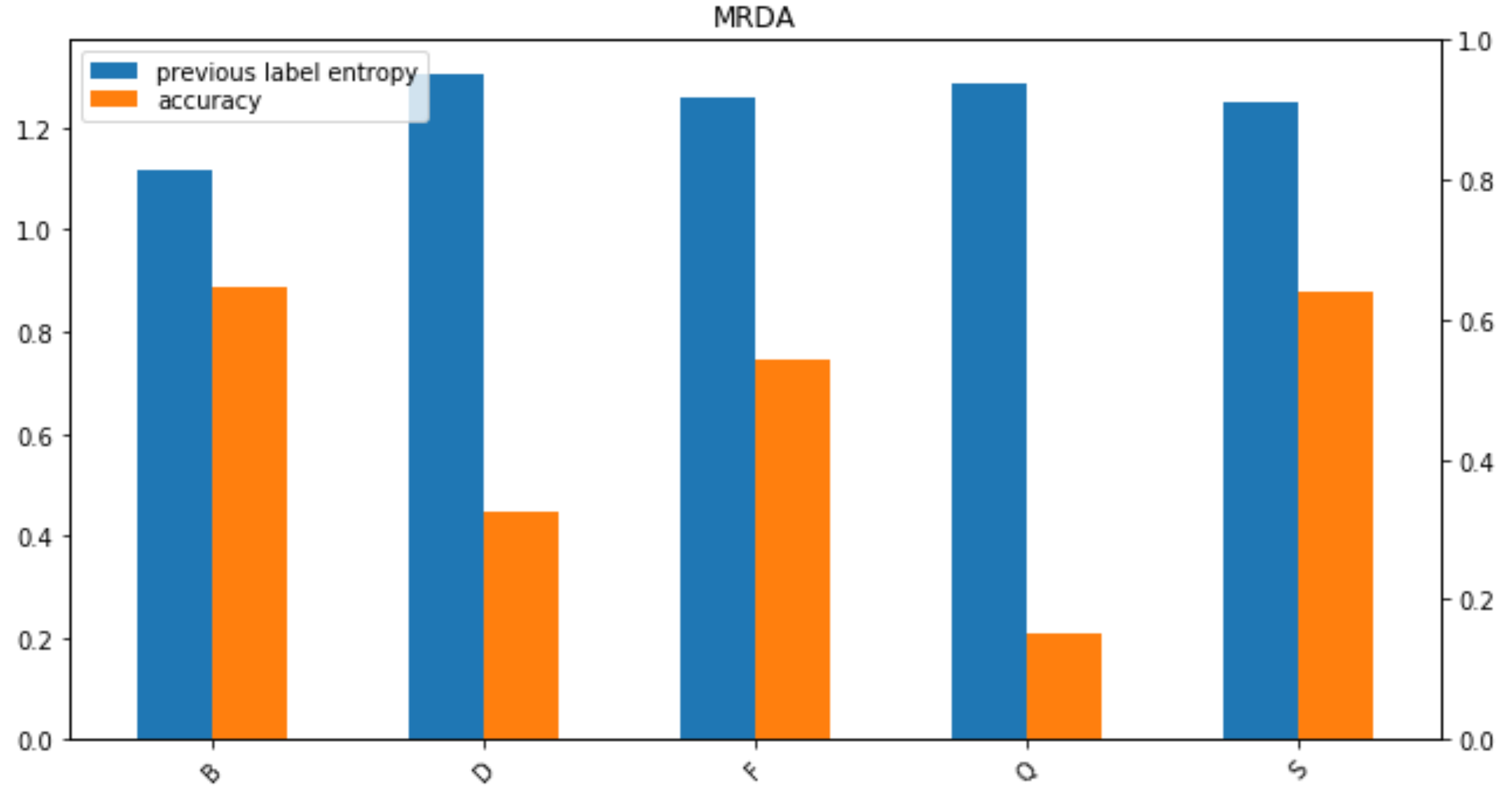}
    \caption{\label{figure3} Previous Label Entropy vs. Accuracy on the MRDA Dataset}
    % \vspace{-0.2cm}
\end{figure}

The improvements that the model is able to make over the other methods are significant, however, the gains on MRDA still fall short of the state-of-the-art by 0.6\%. This can mostly be attributed to the conversation/context lengths and label noise at the conversation level. Conversations in MRDA (1493 utterances on average) are significantly longer than in SwDA  (271  utterances on average). In spite of having nearly 12\% the number of labels (5 vs 43) compared to SwDA, MRDA has 6 times the normalized label entropy in its data. Consequently, due to the noise in label dependencies, and hence, in the inherent conversational structure, the model is not able to yield as big of a gain on the MRDA as it does on the SwDA. Consequently, learning long-range dependencies is a challenge because of noisier and longer path lengths in the network. This is illustrated in Figures \ref{figure2} and \ref{figure3}, which show for every class, the variation between the entropy of the previous label in a conversation, and the accuracy of that class. MRDA was found to have a high negative correlation{\footnote{Pearson's $r$}} (-0.68) between previous label entropy and accuracy, indicating the impact of label noise, which was compounded by longer conversations. On the other hand, SwDA was found to have a low positive correlation (+0.22), which could be compensated by significantly shorter conversations. 
%To test this hypothesis, we evaluated performance of our system on the MRDA corpus by conversation length and found that as the dialogue grew longer, accuracy decreased. 

\begin{table}[t!]
\begin{center}
\small
\begin{tabular}{lll}
\hline \bf Model & \bf SwDA & \bf MRDA\\ \hline
TF-IDF GloVe & 66.5 & 78.7 \\
\citet{kalchbrenner2013recurrent} & 73.9  & - \\
\citet{lee2016sequential} & 73.9 & 84.6 \\
\citet{khanpour2016dialogue} & 75.8 & 86.8 \\
\citet{ji2016latent} & 77.0 & - \\
\citet{shen2016neural} & 72.6 & - \\
\citet{li2016multi} & 79.4 & - \\
\citet{ortega2017neural} & 73.8 & 84.3 \\ 
\citet{tran2017hierarchical} & 74.5 & - \\
\citet{AAAI1816706} & 79.2 & 90.9 \\
\citet{chen2017dialogue} & 81.3 & \bf 91.7 \\ \hline
\bf Our Method & \bf 82.9 & 91.1 \\ \hline
\bf Human Agreement & 84.0 & - \\
\hline
\end{tabular}
\end{center}
\caption{\label{da-classification-results}  DA Classification Accuracy}
\vspace{-0.2cm}
\end{table}

\subsection{Utterance Representation Learning}
\label{utt-rep-learning-section}
%We also investigated the impact that better utterance-level representations can have on the end-task of DA classification,
%One of the primary motivations for this work is the idea whether we can achieve better performance at the task of DA classification by learning better representations for utterances that constitute the conversations. 
One of the primary motivations for this work was to investigate whether one can improve performance by learning better representations for utterances. 
To address this, we retrained our model by replacing the utterance representation learning (utterance-level RNN + context-aware self-attention) component with various sentence representation learning methods (either pre-training them or learning jointly), and feeding them into the conversation-level recurrent layers in the hierarchical model, so that the performance is indicative of the quality of utterance representations. 

There are three main categories of utterance representation learning approaches: (i) the baseline which uses a TF-IDF weighted sum of GloVe word embeddings; (ii)  pre-trained on corpus, where we first learn utterance representations on the corpus using Skip-Thought Vectors \cite{kiros2015skip} and Paragraph Vectors \cite{le2014distributed}, and then use them with the rest of the model; (iii) jointly learned with the DA classification task. Table \ref{utt-rep} describes the performance of different utterance representation learning methods when combined with the overall architecture on both datasets. 

\begin{table}[t!]
\centering
\small 
\begin{tabular}{lll}
\hline \bf Method & \bf SwDA & \bf MRDA \\ \hline
\it{Baseline} \\
TF-IDF GloVe & 66.5 & 78.7 \\
\hline
\it{Pre-trained on Corpus} \\
Skip Thought Vectors & 72.6 & 82.8 \\
Paragraph vectors & 72.5 & 82.6\\
\hline 
\it{Joint Learning} \\
RNN-Encoder & 74.8	& 85.7\\
Bi-RNN-LastState & 76.2 & 85.4 \\
Bi-RNN-MaxPool & 77.6 & 86.7\\
CNN & 76.9 & 84.5 \\
Bi-RNN + Attention & 80.1	& 87.7\\
\hspace{30pt} + Context & 81.8 &	89.2 \\
Bi-RNN + Self-attention & 81.1 & 88.6 \\
\hspace{30pt} + Context & 82.9	& 91.1\\
\hline
\end{tabular}
\caption{\label{utt-rep} Performance of utterance representation methods when integrated with the hierarchical model}
% \vspace{-0.21cm}
\end{table}

Introducing the word-level attention mechanism \cite{yang2016hierarchical} enables the model to learn better representations by attending to more informative words in an utterance, resulting in better performance (Bi-RNN + Attention). The self-attention mechanism (Bi-RNN + Self-attention) leads to even greater overall improvements.  Adding context information (previous recurrent state of the conversation) boosts performance significantly. 

%\section{Results and Discussion}
%\todo{let's make Sec 5 and 6 standalone.  They discuss the experiments, show the results and do the analysis.  Eliminate sec 7. A lot of this section may be redundant already.}
% TODO: PARAPHRASE
% The hierarchical nature of the architecture enables it to represent, interpret, and aggregate structural and contextual dependencies at different levels. Starting from the word level, to the utterance-level, all the way to the conversation-level. Compared to other non-hierarchical models in literature (\citealt{lee2016sequential,kalchbrenner2013recurrent,khanpour2016dialogue} etc.), it is able to model the dependencies internal and external to utterances across a conversation, leading to superior performance. \todo{so is the paper putting a novel method or about fully evaluating a series of heavily overlapping methods that came out in the last year?  This needs to be made clear.  It's kind of hard to follow.}

A notable aspect of our model is how contextual information is leveraged at different levels of the sequence modeling task.  The combination of conversation-level contextual states for utterance-representation learning (+ Context) and a CRF at the conversation level to further inform conversation sequence modeling, leads to a collective performance improvement. 
This is particularly pronounced on the SwDA dataset: the two variants of the context-aware attention models (Bi-RNN + Attention + Context and Bi-RNN + Self-attention + Context) have significant performance gains.%\todo{performance what to what?  which rows should I be looking at?}

% \subsubsection{Richer Embeddings}
% \label{ssec:embedding_results}
% We also tried combining POS-tag embeddings and Sense2Vec to learn word  embeddings, in order to model the structural context, but it didn't perform so well, maybe 

% \begin{figure}
%     \centering
%     \def\svgwidth{\columnwidth}
%     \includegraphics[width=7cm]{attn.png}
%     \caption{\label{figure2} Attention Visualization}
% \end{figure}

\section{Conclusion}
% \todo{mabe sell it harder?}
% We build a hierarchical recurrent neural network, using a bidirectional GRU, with a context-aware self-attention mechanism, followed by a CRF decoder for the task of DA classification. We also experimented with different utterance representation learning methods, and investigated their effectiveness as part of the hierarchical model for the task. 
We developed a model for DA classification with context-aware self-attention, which significantly outperforms earlier models on the commonly-used SwDA dataset and is very close to state-of-the-art on MRDA.
We experimented with different utterance representation learning methods and showed that utterance representations learned at the lower levels can impact the classification performance at the higher level. Employing self-attention, which has not previously been applied to this task, enables the model to learn richer, more effective utterance representations for the task. 

As future work, we would like to experiment with other attention mechanisms such as multi-head attention \cite{vaswani2017attention}, directional self-attention \cite{shen2018disan}, block self-attention \cite{shen2018bidirectional}, or hierarchical attention \cite{yang2016hierarchical}, since they have been shown to address the limitations of vanilla attention and self-attention by either incorporating information from different representation subspaces at different positions to capture both local and long-range context dependencies, encoding temporal order information, or by attending to context dependencies at different levels of granularity.

\section*{Acknowledgements}
The authors would like to thank Dimitris Alikaniotis, Maria Nadejde and Courtney Napoles for their insightful discussions, and the anonymous reviewers for their helpful comments.

\bibliography{naaclhlt2019}
\bibliographystyle{acl_natbib}

\appendix

\section{Supplementary Material}
\label{sec:supplemental}

\subsection{Training \& Hyperparameters}
All hyperparameters were selected by tuning one hyperparameter at a time while keeping the others fixed. Validation splits were used for the tuning process. The final set of hyperparameters were then used to train two different models, one each on SwDA and MRDA training splits. Table \ref{hyperparams} lists the range of values for each parameter that we experimented with, and the final value that was chosen. Dropout was applied to the utterance embeddings. Early stopping was used on the validation set with a patience of 15 epochs. 

\begin{table}[h]
\centering
\small 
\begin{tabular}{lll}
\hline \bf Hyperparams & \bf Range of values & \bf Final value \\ \hline
\hline
Word & GloVe 100$D$ & GloVe 300$D$ + \\ 
Embeddings & GloVe 200$D$ & \hspace{3pt} ELMo 1024$D$ \\
	& GloVe 300$D$ \vspace{-5pt} & \\
	& \hrulefill \\
	& Word2vec 300$D$ & \\
	& Word2vec 200$D$ \vspace{-5pt} & \\
	& \hrulefill \\
	& ELMo 1024$D$ \vspace{-5pt} & \\
	& \hrulefill \\
	& GloVe 300$D$ + \\
	& \hspace{3pt} ELMo 1024$D$ \vspace{-5pt} & \\
	& \hrulefill \\
	& Word2Vec 300$D$ + \\
	& \hspace{3pt} ELMo 1024$D$ & \\
\hline
Sentence GRU & 20 - 300 & 50 \\
Size ($u$) & & \\
\hline
Utterance GRU & 20 - 600 & 100 \\
Size ($k$) & & \\
\hline
% $d_a$ & 100-500 & 200 \\
% \hline
% $r$ & 10-50 & 30 \\
% \hline 
% $$ & 100-3000 & 300 \\
% \hline
Learning Rate & 0.01 - 2.0 & 0.015 \\
\hline
Dropout & 0.1 - 0.8 & 0.3 \\
\hline
Optimizer & SGD, & Adam \\
 & RMSProp, & \\
 & Adam & \\
\hline
\end{tabular}
\caption{\label{hyperparams} Hyperparameter space and tuned values}
\end{table}

% Nonetheless, supplementary material should be supplementary (rather
% than central) to the paper. {\bf Submissions that misuse the supplementary 
% material may be rejected without review.}
% Supplementary material may include explanations or details
% of proofs or derivations that do not fit into the paper, lists of
% features or feature templates, sample inputs and outputs for a system,
% pseudo-code or source code, and data. (Source code and data should
% be separate uploads, rather than part of the paper).

% The paper should not rely on the supplementary material: while the paper
% may refer to and cite the supplementary material and the supplementary material will be available to the
% reviewers, they will not be asked to review the
% supplementary material.

\end{document}

%% file: Model.pdf_tex
%% Creator: Inkscape inkscape 0.91, www.inkscape.org
%% Accompanies image file 'Model.pdf' (pdf, eps, ps)
\begingroup%
  \makeatletter%
  \providecommand\color[2][]{%
    \errmessage{(Inkscape) Color is used for the text in Inkscape, but the package 'color.sty' is not loaded}%
    \renewcommand\color[2][]{}%
  }%
  \providecommand\transparent[1]{%
    \errmessage{(Inkscape) Transparency is used (non-zero) for the text in Inkscape, but the package 'transparent.sty' is not loaded}%
    \renewcommand\transparent[1]{}%
  }%
  \providecommand\rotatebox[2]{#2}%
  \ifx\svgwidth\undefined%
    \setlength{\unitlength}{1050.84931641bp}%
    \ifx\svgscale\undefined%
      \relax%
    \else%
      \setlength{\unitlength}{\unitlength * \real{\svgscale}}%
    \fi%
  \else%
    \setlength{\unitlength}{\svgwidth}%
  \fi%
  \global\let\svgwidth\undefined%
  \global\let\svgscale\undefined%
  \makeatother%
  \begin{picture}(1.2,1.06523822)%
    \put(0,-0.32){\includegraphics[width=\unitlength,page=1]{Model.pdf}}%
  \end{picture}%
\endgroup%